\documentclass[letterpaper]{article} 
\usepackage[preprint]{aaai2027}  
\usepackage[hyphens]{url}  
\usepackage{graphicx} 
\usepackage{natbib}  
\usepackage{caption} 
\usepackage{algorithm}
\usepackage{algorithmic}

\usepackage{newfloat}
\usepackage{listings}
\DeclareCaptionStyle{ruled}{labelfont=normalfont,labelsep=colon,strut=off} 
\floatstyle{ruled}
\newfloat{listing}{tb}{lst}{}
\floatname{listing}{Listing}

\usepackage{booktabs}
\usepackage{amssymb}
\usepackage{amsmath}
\usepackage{pifont} 
\usepackage{array} 
\newcommand{\cmark}{\ding{51}}
\newcommand{\xmark}{\ding{55}}
\title{Quick on the Uptake: Eliciting Implicit Intents from \\ Human Demonstrations for Personalized Mobile-Use Agents}

\author{
    Zheng Wu\textsuperscript{\rm 1},
    Heyuan Huang\textsuperscript{\rm 2},
    Yanjia Yang\textsuperscript{\rm 1},
    Yuanyi Song\textsuperscript{\rm 1},
    Xingyu Lou\textsuperscript{\rm 2},\\
    Weiwen Liu\textsuperscript{\rm 1},
    Weinan Zhang\textsuperscript{\rm 1},
    Jun Wang\textsuperscript{\rm 2},
    Zhuosheng Zhang\textsuperscript{\rm 1}\corresponding
}
\affiliations{
    \textsuperscript{\rm 1}School of Computer Science, Shanghai Jiao Tong University\\
    \textsuperscript{\rm 2}OPPO Research Institute\\
    \{wzh815918208,zhangzs\}@sjtu.edu.cn \quad junwang.lu@gmail.com
}

\begin{document}

\maketitle

\begin{abstract}
As multimodal large language models advance rapidly, the automation of mobile tasks has become increasingly feasible through the use of mobile-use agents that mimic human interactions from graphical user interfaces. 
To further enhance mobile-use agents, previous studies employ demonstration learning to improve mobile-use agents from human demonstrations. However, these methods focus solely on the explicit intention flows of humans (e.g., step sequences) while neglecting implicit intention flows (e.g., personal preferences), which makes it difficult to construct personalized mobile-use agents. 
In this work, to evaluate the Intention Alignment Rate between mobile-use agents and humans, we first collect MobileIAR, a dataset covering 40 English- and Chinese-speaking users and containing human-intent-aligned actions and ground-truth actions. 
This enables a comprehensive assessment of the agents' understanding of human intent. 
Then we propose IFRAgent, a framework built upon Intention Flow Recognition from human demonstrations. 
IFRAgent analyzes explicit intention flows from human demonstrations to construct a query-level vector library of standard operating procedures (SOP), and analyzes implicit intention flows to build a user-level habit repository. 
IFRAgent then leverages a SOP extractor combined with retrieval-augmented generation and a query rewriter trained through knowledge distillation to generate personalized queries and SOPs from a raw ambiguous query, enhancing the alignment between mobile-use agents and human intent. 
Experimental results demonstrate that IFRAgent consistently outperforms baselines on every model, achieving an average absolute improvement of 9.88\% (53.33\% relative) in human intention alignment rate and an average absolute improvement of 8.26\% (32.40\% relative) in step completion rate.

\end{abstract}

\section{Introduction}
\label{sec:intro}

\begin{figure}[t]
\centering
\includegraphics[width=0.48\textwidth]{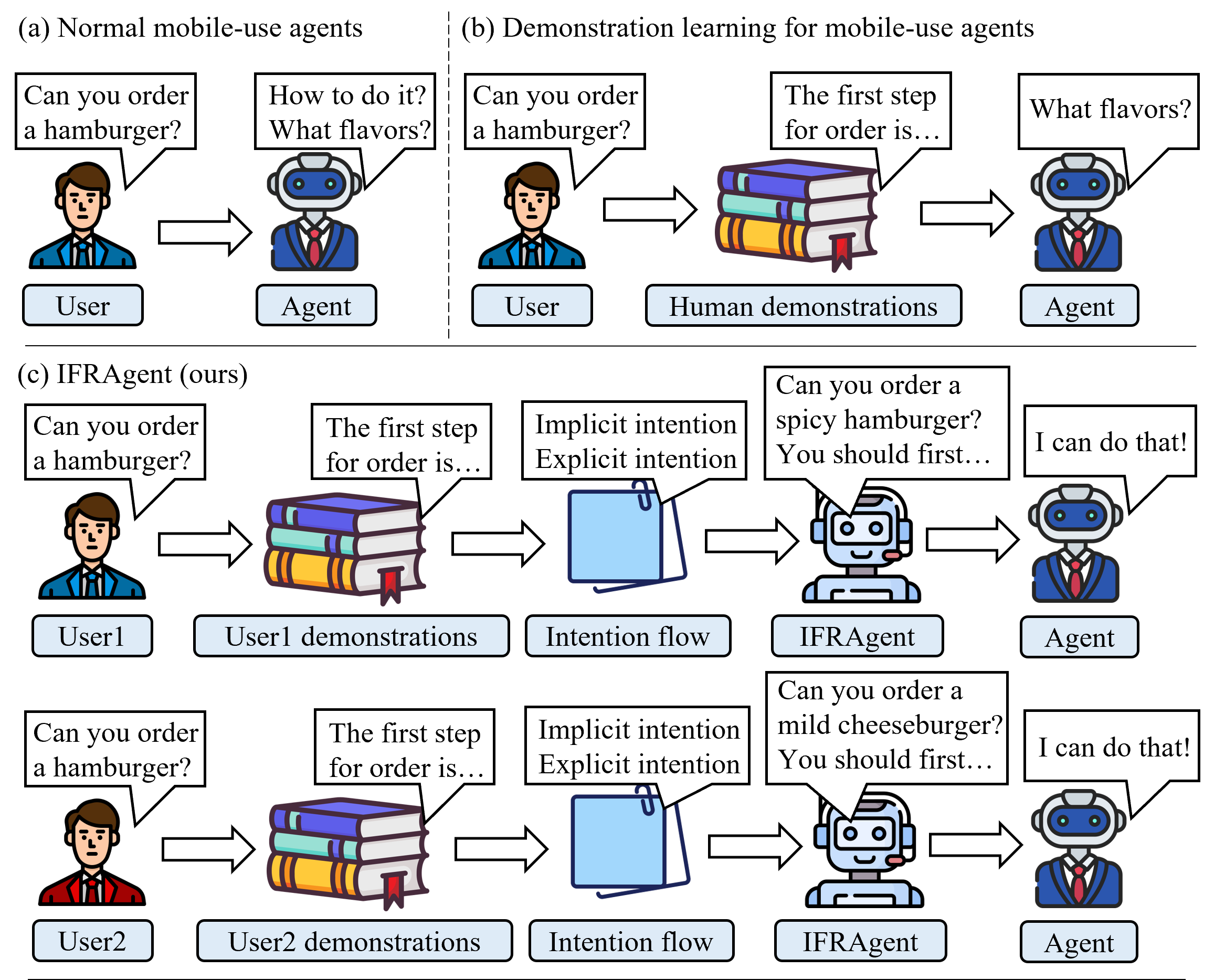}
\caption{Unlike existing demonstration learning methods that capture explicit intention flow, IFRAgent also models implicit intention flow (e.g., preferences) for personalization.}
\label{intro}
\end{figure}

As multimodal large language models advance rapidly~\cite{zhang-etal-2024-mm, yin2024survey,chen2026trace}, mobile task automation has become increasingly feasible through mobile-use agents that mimic human GUI interactions (e.g., clicking, scrolling)~\cite{zhang2024large, liu2025llm,tang2025surveymllmbasedguiagents}, with some works~\cite{liu2025learnact,verma2024adaptagent} further employing demonstration learning to let these agents learn how to complete tasks from human demonstrations.

However, as shown in Figure~\ref{intro}, existing demonstration learning methods for mobile-use agents focus solely on explicit human intention flows (e.g., operational logic, step sequences) to help mobile-use agents learn how to complete tasks. 
Moreover, user instructions in the real world are often ambiguous~\cite{cheng2025navi} and user-specific, requiring mobile-use agents to understand the implicit intention flows of humans in order to align with human intentions. 

There are two challenges for mobile-use agents to align with human intentions: (i) There is a lack of datasets or benchmarks that can assess the alignment level between mobile-use agents and human intentions. 
(ii) Fine-tuning a separate mobile-use agent for each user to create user-specific mobile-use agents is impractical.

For challenge (i), we first collect \textbf{MobileIAR}. This user-specific dataset supports both English and Chinese language users, designed to assess the \textbf{I}ntention \textbf{A}lignment \textbf{R}ate between mobile-use agents and humans. 
The dataset covers 16 apps and involves 40 users (9 human annotators and 31 synthesized virtual users), contributing a total of 4,194 instructions spanning seven daily scenarios. Moreover, the dataset provides both the human-intent-aligned actions and ground-truth action lists, enabling a comprehensive assessment of the alignment level between mobile-use agents and human intentions.
Even for the same query, different users may take different human-intent-aligned actions according to their own preferences, while a broader ground-truth action list also covers other actions that can still fulfill the query; this distinction allows MobileIAR to jointly measure both intent alignment and general task completion.

For challenge (ii), we propose \textbf{IFRAgent}, a plug-and-play framework built upon \textbf{I}ntention \textbf{F}low \textbf{R}ecognition from human demonstrations.
IFRAgent first extracts both explicit and implicit intention flows from human demonstrations, and then leverages retrieval-augmented generation~\cite{lewis2020retrieval} together with a knowledge-distilled query rewriter to rewrite a user's ambiguous query into a personalized query and SOP at deployment time.
This enables mobile-use agents to better align with human intentions, while also improving their general task completion capability to a certain extent.

Extensive experiments spanning diverse mobile-use agents (supporting multiple user languages and constructed with varying methods) demonstrate that IFRAgent consistently outperforms baseline methods on every model and every metric, by an average of 9.88\% (53.33\% relative improvement) in intent alignment rate and an average of 8.26\% (32.40\% relative improvement) in task completion rate.
We also find that general-domain models (e.g., Qwen2.5-VL-7B, Qwen3-VL-8B, GPT-4o) demonstrate more significant improvements compared to specialized mobile-use base models.

We further validate IFRAgent's capability and generalizability through an ablation study, cross-dataset tests, comparative studies with other methods, and scale analysis.

In summary, we make three key contributions: 

(i) we contribute and open-source MobileIAR, the first benchmark for user-specific intent alignment testing in the field of mobile-use agents.

(ii) we propose IFRAgent, a plug-and-play framework that leverages both explicit and implicit intention flows from human demonstrations to enhance mobile-use agents' task completion capability and user-specific intent alignment.

(iii) extensive experiments show that IFRAgent consistently improves mobile-use agents over baseline methods.

\begin{table}[!t]
    \centering
    \footnotesize
    \renewcommand\arraystretch{1.1}
    \setlength{\tabcolsep}{2.7pt}
    \begin{tabular}{p{2.2cm}ccc>{\centering\arraybackslash}p{0.58cm}>{\centering\arraybackslash}p{0.58cm}>{\centering\arraybackslash}p{0.58cm}>{\centering\arraybackslash}p{0.58cm}>{\centering\arraybackslash}p{0.58cm}}
    \toprule
    \textbf{Dataset} & \textbf{\# Inst.} & \textbf{\# Apps} &\textbf{\# Step} &\textbf{HL} & \textbf{GT} & \textbf{FS} & \textbf{US}\\
    \midrule
    PixelHelp & 187 & 4 & 4.2 & \cmark & \cmark & \xmark & \xmark\\
    MoTIF & 276 & 125 & 4.5 & \cmark & \cmark & \xmark & \xmark\\
    UIBert & 16,660 & - & 1& \xmark & \cmark & \xmark & \xmark\\
    Meta-GUI & 1,125 & 11 & 15 & \cmark & \cmark & \xmark & \xmark \\
    UGIF & 523 & 12 & 6.3 &\cmark & \cmark & \xmark & \xmark \\
    AITW & 30,378 & 357 & 6.5 & \cmark & \cmark & \xmark & \xmark\\
    AITZ & 2,504 & 70 & 7.5 & \cmark & \cmark & \xmark & \xmark\\
    AndroidControl & 15,283 & 833 & 4.8 & \cmark & \cmark & \xmark & \xmark\\
    AMEX & 2,946 & 110  & 12.8 & \cmark & \cmark & \xmark & \xmark \\
    OS-Kairos & 1000 & 12 & 5.1& \cmark & \cmark & \xmark & \xmark \\
    MobileAgentBench & 100 & 10 &- & \cmark & \xmark & \xmark & \xmark\\
    AppAgent & 50 & 10&-  & \cmark& \xmark & \xmark & \xmark\\
    LlamaTouch & 496 & 57 & 7.0 & \cmark & \cmark & \xmark & \xmark \\
    AndroidWorld & 116 & 20 & - & \cmark & \xmark & \xmark & \xmark\\
    AndroidLab & 138 & 9 & 8.5  & \cmark& \xmark & \xmark & \xmark\\
    LearnGUI & 2353 & 73 & 13.2 & \cmark & \cmark & \cmark & \xmark\\
    \midrule
    \textbf{MobileIAR} & \textbf{4,194} & \textbf{16}  & \textbf{7.5} & \textbf{\cmark} & \textbf{\cmark} & \textbf{\cmark} & \textbf{\cmark}\\
    \bottomrule
    \end{tabular}
    \caption{Comparison of datasets and environments for evaluating mobile-use agents. \# Inst./\# Apps/\# Step: instruction/app/average-step counts; HL: high-level instructions; GT: ground-truth trajectories; FS: few-shot support; US: user-specific demonstrations. Data partially sourced from~\citet{liu2025learnact}.}
    \label{tab:datasets}
\end{table}

\begin{figure*}[t]
\centering
\includegraphics[width=\textwidth]{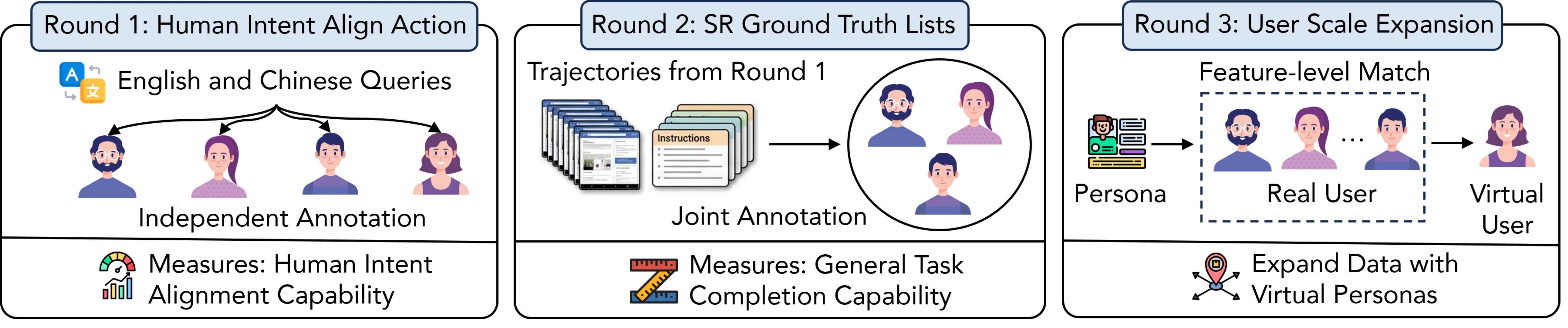}
\caption{Construction of MobileIAR: Round 1 collects human-intent-aligned actions from English/Chinese annotators; Round 2 jointly labels SR ground-truth lists; Round 3 expands the 9 seed annotators into 40 users via synthesized virtual personas.}
\label{bench}
\end{figure*}

\section{Related work}
\subsection{Mobile-use agents for mobile automation}
Mobile-use agents~\cite{wang2024gui,hu-etal-2025-os} automate mobile tasks by simulating human GUI interactions (e.g., clicking, scrolling), and are typically built either from open-source (M)LLMs post-trained via reinforcement learning~\cite{zhang2023you,wuatlas,qin2025ui,wang2024distrl,lu2025ui} or by leveraging the general-domain knowledge of closed-source models~\cite{jiang2025appagentx,wang2025mobileE,li2025mobileuse}.
As shown in Table~\ref{tab:datasets}, existing benchmarks for evaluating mobile-use agents rely on either static images~\cite{shaw2023pixels,rawles2024androidinthewild,zhang2024android} or dynamic environments~\cite{wang2024mobile,rawles2024androidworld,xu2024androidlab}, but all only assess task completion, lacking a benchmark for the alignment level between mobile-use agents and human intentions.

\subsection{Demonstration learning for mobile-use}
Demonstration learning~\cite{correia2024survey}, encompassing imitation learning~\cite{rybski2007interactive} and inverse reinforcement learning~\cite{ng2000algorithms}, enhances agent capability by observing human demonstrations and has been widely applied in the robot domain~\cite{argall2009survey}.
For mobile-use and computer-use agents, prior work leverages few-shot demonstrations~\cite{liu2025learnact,verma2024adaptagent} or extracts references and training data from videos~\cite{zhang2026tongui,jung2026tvagent,wang2025mobile}, but neglects implicit intention flows such as user behavioral preferences.
IFRAgent comprehensively analyzes both explicit and implicit intention flows, establishing a more user-intent-aligned paradigm for mobile-use agent demonstration learning.

\section{MobileIAR dataset collection}
As shown in Figure~\ref{bench}, we construct MobileIAR through three rounds: two rounds of crowdsourced annotation followed by a round of user scale expansion that synthesizes virtual users.

\subsection{Round 1: Human Intent Align Action}
We first collect trajectories for both English-speaking and Chinese-speaking users across seven daily-life scenarios through a crowdsourcing approach, resulting in a total of 945 instructions and 7,310 screenshots.
The annotators cover a diverse group of participants, including men and women from different age groups.
In this round, annotators independently collect trajectories on their own real mobile devices, and are encouraged to choose completion methods that highlight their personalized preferences for the same instruction.
The results of this round are used as the human-intent-aligned action, which measures the ability of the mobile-use agent to align with the user's intent.

\subsection{Round 2: SR Ground Truth Lists}
The second round of annotation builds upon the first round, aiming to allow MobileIAR to assess not only the mobile-use agent’s intent alignment capability but also its general competence.
Since the human-intent-aligned action is not always the only correct action in some cases, and often the agent's execution path may have multiple routes,
we form groups of three annotators for each instruction and ask them to jointly label, from scratch, all possible correct alternative actions for every step.
The union of these labeled actions is then taken as the step-wise success rate (SR) ground-truth action list, which measures the mobile-use agent's general task completion ability.


\subsection{Round 3: User Scale Expansion}
Since scaling the number of distinct users purely through crowdsourcing is costly and slow, we further expand the 9 manually annotated seed users into a total of 40 users by synthesizing 31 additional virtual users through an LLM-driven persona synthesis pipeline. For each virtual user, an LLM first generates a demographically-constrained profile (nationality, age, gender, occupation, city, and per-domain app/behavior preferences); the LLM then matches each domain and app of this profile to the most behaviorally similar seed user among the 9 annotators, and the virtual user's habit repository, SOP library, support-set trajectories, and test instructions are assembled by reusing and personalizedly rewriting the corresponding real trajectories of the matched seed users. Notably, this procedure only recombines and rewrites existing human-collected trajectories rather than fabricating new screenshots, so all reported metrics remain grounded in real human demonstrations.
Validation with 15 sampled virtual users and three independent annotators (1–5 scale) yielded an average rating of 4.3, confirming that the rewriting process reliably preserves persona-consistent behavior.

\section{IFRAgent Framework}

\begin{figure*}[h]
\centering
\includegraphics[width=\textwidth]{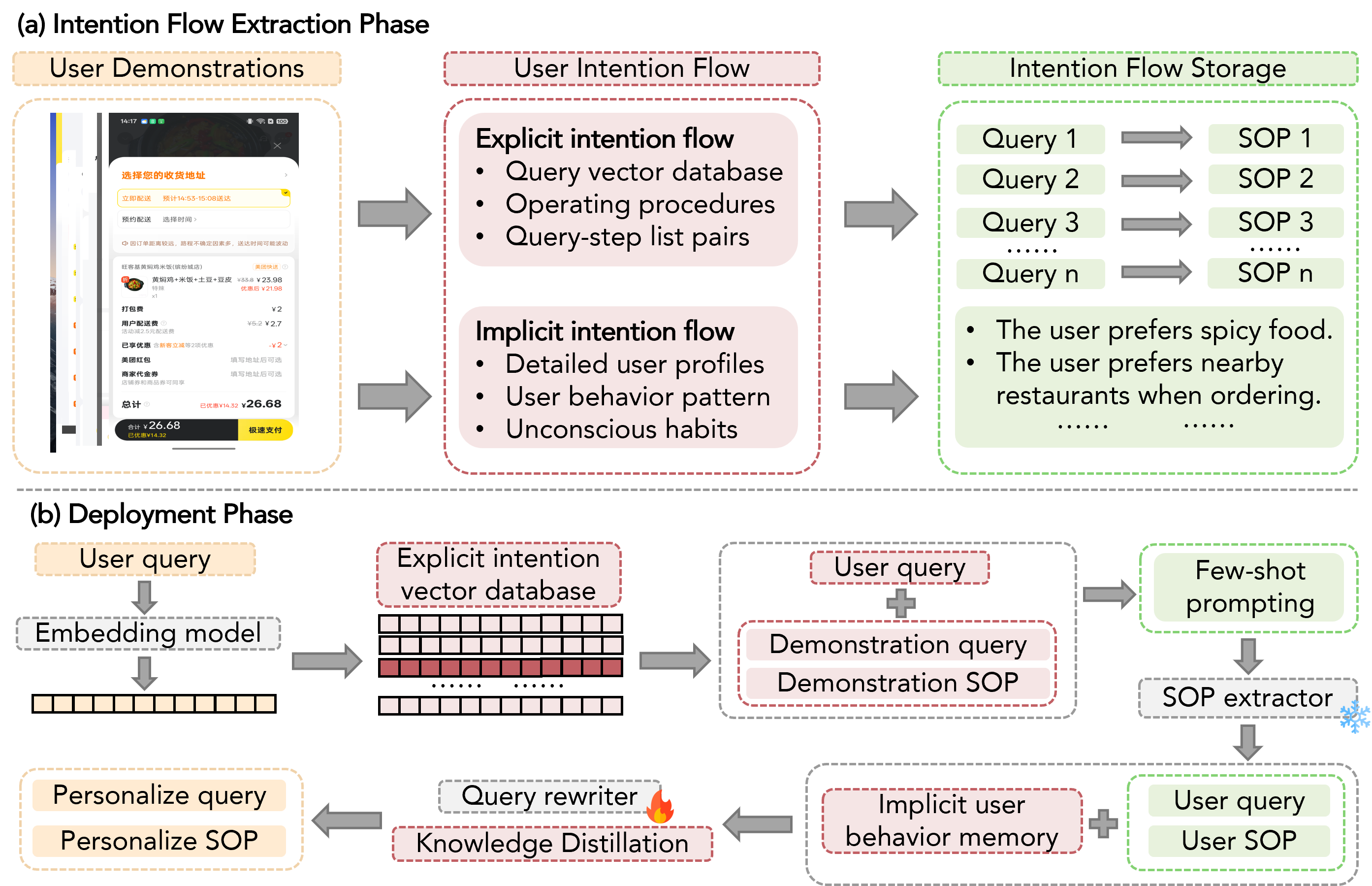}
\caption{The workflow pipeline of IFRAgent. Above the dashed line, the intention flow extraction phase derives SOPs and user profiles from trajectories; below it, the deployment phase uses this information to rewrite personalized queries and SOPs.}
\label{pipeline}
\end{figure*}

We propose IFRAgent, a framework that enhances intent alignment between mobile-use agents and humans; Figure~\ref{pipeline} shows an overview of the pipeline and the trainable scheme via knowledge distillation, introduced below.

\subsection{IFRAgent Pipeline}
The IFRAgent pipeline can be divided into 2 phases: the intention flow extraction phase and the deployment phase. 

\subsubsection{Intention Flow Extraction Phase}




In the intention flow extraction phase, IFRAgent collects human demonstrations and analyzes these human demonstrations to extract the implicit intention flow and explicit intention flow of humans. 
For each user $u_i \in U = \{u_1, u_2, \dots, u_n\}$, we first collect human demonstrations comprising a set of queries $Q_i = \{q_1, q_2, \dots, q_k\}$ and initialize an empty user-level habit repository $h_i$. 
Each query $q_j \in Q_i$ is accompanied by a sequence of operation trajectory screenshots $S(u_i, q_j) = \{s_1, s_2, \dots, s_p\}$ provided by the user.  

Given the tuple $(q_j, S(u_i, q_j))$, we first process it through an explicit intention flow agent $A_e$ to extract a SOP: $p_j = A_e(q_j, S(u_i, q_j))$.

Concurrently, the query $q_j$ is encoded into a latent representation $\mathbf{l}_j$ using an embedding model $\phi$, such that $\mathbf{l}_j = \phi(q_j)$. The pair $(\mathbf{l}_j, p_j)$ is stored for $u_i$ to facilitate retrieval during deployment.  

Simultaneously, the tuple is processed through an implicit intention flow agent $A_i$ that incrementally updates the habit repository: $h_i \gets h_i \cup \{ A_i(h_i, q_j, S(u_i, q_j)) \}$, where $A_i$ learns latent behavioral patterns from interaction sequences. This dual-processing framework iterates over all queries in $Q_i$ until all human demonstrations are consumed, resulting in a comprehensive habit repository $h_i$ and retrievable explicit SOPs $\{(\mathbf{l}_j, p_j)\}$ for each user.

\begin{table*}[t]
\centering
\begin{tabular}{lp{1.5cm}p{1.5cm}p{1.5cm}p{1.5cm}p{1.5cm}p{1.5cm}}
\toprule
& \multicolumn{3}{c}{English users} & \multicolumn{3}{c}{Chinese users} \\
\cmidrule(lr){2-4} \cmidrule(lr){5-7}
\textbf{~~~~~~~~~~~~~Model} & SR(\%)$\uparrow$ & Type(\%)$\uparrow$ & IAR(\%)$\uparrow$ & SR(\%)$\uparrow$ & Type(\%)$\uparrow$ & IAR(\%)$\uparrow$ \\
\midrule

\multicolumn{7}{c}{\textbf{Open-source based mobile-use agents}} \\
\midrule
OS-Atlas-7B-Pro & 39.81 & 74.40 & 40.61 & 42.41 & 74.48 & 32.15 \\
~~~~~~~~~~~~~~~~~~+ IFRAgent & 47.72$_{7.91\uparrow}$ & 81.46$_{7.06\uparrow}$ & 47.99$_{7.38\uparrow}$ & 55.15$_{12.73\uparrow}$ & 82.75$_{8.27\uparrow}$ & 47.71$_{15.55\uparrow}$ \\
\midrule
UI-TARS-7B-SFT & 40.35 & 65.61 & 38.43 & 44.40 & 74.02 & 36.66 \\
~~~~~~~~~~~~~~~~~~+ IFRAgent & 45.95$_{5.60\uparrow}$ & 79.79$_{14.18\uparrow}$ & 45.66$_{7.22\uparrow}$ & 57.23$_{12.83\uparrow}$ & 81.44$_{7.42\uparrow}$ & 51.62$_{14.97\uparrow}$ \\
\midrule
UI-TARS-7B-DPO & 42.16 & 71.84 & 33.10 & 44.33 & 70.50 & 36.86 \\
~~~~~~~~~~~~~~~~~~+ IFRAgent & 48.44$_{6.28\uparrow}$ & 72.51$_{0.67\uparrow}$ & 40.10$_{7.00\uparrow}$ & 48.24$_{3.91\uparrow}$ & 76.51$_{6.01\uparrow}$ & 44.89$_{8.03\uparrow}$ \\
\midrule
UI-TARS-1.5-7B & 41.18 & 72.95 & 34.17 & 46.96 & 73.69 & 40.29 \\
~~~~~~~~~~~~~~~~~~+ IFRAgent & 44.76$_{3.58\uparrow}$ & 79.50$_{6.55\uparrow}$ & 43.56$_{9.39\uparrow}$ & 52.26$_{5.29\uparrow}$ & 82.81$_{9.11\uparrow}$ & 48.41$_{8.12\uparrow}$ \\
\midrule
Qwen2.5-VL-7B & 15.91 & 14.13 & 8.99 & 18.57 & 21.85 & 17.01 \\
~~~~~~~~~~~~~~~~~~+ IFRAgent & 32.27$_{16.36\uparrow}$ & 41.06$_{26.93\uparrow}$ & 30.68$_{21.69\uparrow}$ & 38.91$_{20.34\uparrow}$ & 49.95$_{28.10\uparrow}$ & 40.74$_{23.73\uparrow}$ \\
\midrule
GUI-owl-7B & 43.32 & 82.62 & 38.21 & 40.92 & 68.35 & 34.17 \\
~~~~~~~~~~~~~~~~~~+ IFRAgent & 53.87$_{10.55\uparrow}$ & 87.21$_{4.58\uparrow}$ & 45.83$_{7.62\uparrow}$ & 47.68$_{6.75\uparrow}$ & 75.28$_{6.93\uparrow}$ & 42.11$_{7.94\uparrow}$ \\
\midrule
Qwen3-VL-8B & 42.43 & 75.90 & 37.47 & 37.81 & 63.55 & 30.63 \\
~~~~~~~~~~~~~~~~~~+ IFRAgent & 51.81$_{9.37\uparrow}$ & 85.54$_{9.64\uparrow}$ & 46.65$_{9.18\uparrow}$ & 45.58$_{7.77\uparrow}$ & 71.31$_{7.76\uparrow}$ & 37.92$_{7.29\uparrow}$ \\
\midrule

\multicolumn{7}{c}{\textbf{Close-source based mobile-use agents}} \\
\midrule
GPT-4o + OCR model & 36.88 & 76.79 & 30.76 & 37.15 & 78.00 & 31.06 \\
~~~~~~~~~~~~~~~~~~+ IFRAgent & 44.53$_{7.66\uparrow}$ & 79.94$_{3.16\uparrow}$ & 38.54$_{7.78\uparrow}$ & 49.34$_{12.19\uparrow}$ & 81.64$_{3.64\uparrow}$ & 46.27$_{15.21\uparrow}$ \\
\midrule
GLM-4v + OCR model & 6.43 & 56.35 & 2.56 & 4.91 & 76.61 & 3.28 \\
~~~~~~~~~~~~~~~~~~+ IFRAgent & 9.60$_{3.17\uparrow}$  & 77.20$_{20.85\uparrow}$  & 5.03$_{2.47\uparrow}$  & 7.97$_{3.06\uparrow}$  & 79.90$_{3.29\uparrow}$  & 7.38$_{4.10\uparrow}$  \\
\midrule
Qwen-VL-max + OCR model & 17.96 & 81.09 & 16.08 & 21.70 & 84.68 & 15.20 \\
~~~~~~~~~~~~~~~~~~+ IFRAgent & 23.59$_{5.64\uparrow}$ & 86.65$_{5.56\uparrow}$ & 22.27$_{6.19\uparrow}$ & 25.85$_{4.15\uparrow}$ & 90.36$_{5.68\uparrow}$ & 21.85$_{6.66\uparrow}$ \\

\bottomrule
\end{tabular}
\caption{Performance of open- and closed-source mobile-use agents with IFRAgent, which consistently improves all metrics.}
\label{mainresult}
\end{table*}

\subsubsection{Deployment Phase}




In the deployment phase, when processing a user query $q$ from user $u_i$, we first encode it into vector $\mathbf{l}$ using the same embedding model $\phi$ as employed in the intention flow extraction phase (where $\mathbf{l}$ = $\phi(q)$).
Then we match $\mathbf{l}$ against each explicit intention flow $(\mathbf{l}_j, p_j)$ of the user $u_i$ for RAG. 

When the similarity exceeds a threshold $\tau$, we obtain the most similar query $q'$ and its corresponding SOP $p'$:

\begin{align*}
(q', p') =
\begin{cases}
(q_j, p_j) & \text{if } \exists j = \arg\max_k \text{sim}(\mathbf{l}, \mathbf{l}_k) \\
           & \quad \land\ \text{sim}(\mathbf{l}, \mathbf{l}_k) > \tau, \\
\varnothing & \text{otherwise.}
\end{cases}
\end{align*}

The query $q'$, its SOP $p'$, and the query $q$ are used together as the prompt for few-shot learning and then fed into the SOP Extractor $\mathcal{E}$ to obtain the SOP $p$ corresponding to $q$: $p = \mathcal{E}(q, (q', p'))$.
Next, the query $q$ and its corresponding SOP $p$ are combined with the user habit repository $h_i$ of user $u_i$ as input to the query rewriter $\mathcal{W}$ to generate a rewritten personalized query $\hat{q}$ and SOP $\hat{p}$ that align user-specific intention: $(\hat{q}, \hat{p}) = \mathcal{W}(q, p, h_i)$.
Finally, the rewritten query $\hat{q}$, the rewritten SOP $\hat{p}$, and the current screenshot $s$ are provided as input to the mobile-use agent $\mathcal{F}$ to obtain the action $a$: $a = \mathcal{F}(\hat{q}, \hat{p}, s)$.

\subsection{Training via Knowledge Distillation}

The SOP Extractor $\mathcal{E}$ and the query rewriter $\mathcal{W}$ are two key components of the IFRAgent during the deployment phase. 
Both $\mathcal{E}$ and $\mathcal{W}$ are models deployable on the edge side, thus lacking general knowledge about mobile operations. 
Since $\mathcal{E}$ has already compensated for the lack of general knowledge through few-shot learning with RAG, only $\mathcal{W}$ requires supervised fine-tuning to unleash its potential in query rewriting.

To warm up the trainable scheme query rewriter $\mathcal{W}$, we employ a data distillation method, allowing $\mathcal{W}$ to learn the query and SOP rewriting logic from large-scale models with strong general knowledge (e.g., GPT-4o). 
For the support dataset of MobileIAR, we used GPT-4o to replace the query rewriter $\mathcal{W}$ during the deployment phase, obtaining personalized rewritten queries \( \hat{q} \) and personalized rewritten SOPs \( \hat{p} \) for each query. Then, using these \( \hat{q} \) and \( \hat{p} \) as labels, we performed supervised fine-tuning on Qwen3-4B to warm up the query rewriter $\mathcal{W}$, enabling it to learn how to rewrite queries and SOPs using personalized user information. 
The training objective can be formulated as: 
\begin{align*}
\mathcal{L}_{\text{SFT}} = \mathbb{E}_{(q, p, h_i, \hat{q}, \hat{p}) \sim \mathcal{D}} \left[ \mathcal{L}\left( \mathcal{W}(q, p, h_i), (\hat{q}, \hat{p}) \right) \right].
\end{align*}

After the warm-up training, $\mathcal{W}$ acquires the capability to rewrite queries and SOPs in a user-specific manner.




\section{Experiments}

\subsection{Experiments Setup}
\subsubsection{Baseline}
Mobile-use agents can be categorized by their construction methods into open-source and closed-source agents. 
We selected seven state-of-the-art open-source-based mobile-use agents and three closed-source-based mobile-use agents as baselines to experimentally validate the extent to which IFRAgent enhances mobile-use agents. 
The open-source based mobile-use agents include OS-Atlas-7B-Pro~\cite{wuatlas}, UI-TARS-7B-SFT, UI-TARS-7B-DPO~\cite{qin2025ui}, UI-TARS-1.5-7B, Qwen2.5-VL-7B~\cite{Qwen2.5-VL}, GUI-owl-7B~\cite{ye2025mobile} and Qwen3-VL-8B~\cite{Qwen3-VL} while the close-source based mobile-use agents include GPT-4o~\cite{openai2023gpt4}, GLM-4v~\cite{glm2024chatglm}, and Qwen-VL-max~\cite{Qwen-VL}. 
Since GPT-4o, GLM-4v, and Qwen-VL-max inherently lack the capability to predict coordinates, we incorporated an OCR model composed of ResNet18~\cite{he2016deep} and ConvNeXt-Tiny~\cite{liu2022convnet} to assist them in localization.
Our main experiments were conducted on our collected dataset, MobileIAR. 
MobileIAR is the first benchmark for user-specific intent alignment testing in the field of mobile-use agents.
In this experiment, both the implicit intention flow agent and the explicit intention flow agent are based on GPT-4o, while the SOP extractor and query rewriter are based on Qwen3-4B.

\subsubsection{Metric}
We consider two types of metrics: one measures the task completion capability of mobile-use agents, and the other measures the alignment level between mobile-use agents and human intentions. 
Following existing work on mobile-use agents~\cite{zhang2023you,ma2024comprehensive,wuatlas,qin2025ui,cheng2025kairos}, we report the SR and action type accuracy (Type) to assess task completion. To measure the alignment level between mobile-use agents and human intentions, we report the intention alignment rate (IAR).
In the MobileIAR dataset, we provide human-intent-aligned actions and ground-truth action lists. The human-intent-aligned action is the single most aligned action with the user's intent at each step, while ground-truth action lists are a set of possible actions that could help fulfill the user's query in the current frame. For metric calculations, SR and Type consider the mobile-use agent's action correct if it matches any of the ground-truth actions at the current step. In contrast, IAR requires the agent's action to exactly match the human-intent-aligned action to be counted as correct. 
All evaluations are conducted with do\_sample=False (greedy decoding), which yields deterministic outputs; we therefore report results from a single run.

\subsection{Main Results}

As shown in Table~\ref{mainresult}, we conducted extensive experiments on mobile-use agents constructed using ten different methods, yielding the following key findings.

(i) IFRAgent consistently improves every mobile-use agent on every metric, for both English and Chinese users, with absolute (relative) improvements of SR by 8.26\% (32.40\%) and IAR by 9.88\% (53.33\%), confirming its effectiveness in both general task completion and user-specific intent alignment.

(ii) General-domain models (Qwen2.5-VL-7B, Qwen3-VL-8B, GPT-4o) benefit more than specialized mobile-use agent base models like UI-TARS, since post-training on GUI operations tends to erode the general world knowledge useful for intention recognition.

(iii) GLM-4v and Qwen-VL-max show a much lower SR than Type, indicating that they often infer the correct action type but fail to use the OCR-provided localization; since IFRAgent does not supply localization information, its improvement on these two agents is comparatively smaller.

(iv) Mobile-use agents generally perform better for Chinese users, as Chinese app designs are more standardized and models such as OS-Atlas-7B-Pro and UI-TARS are more extensively trained on Chinese-environment data. Overall, these findings suggest that general language understanding sets the lower bound of a mobile-use agent's personalized capability, while its mobile-domain execution capability sets the upper bound.

\section{Further Analysis}
In this section, we assess the contribution of individual components, evaluate generalizability across datasets, analyze scaling behavior, and measure computational efficiency.
\subsection{Ablation Study}
\begin{figure}[t]
\includegraphics[width=0.5\textwidth]{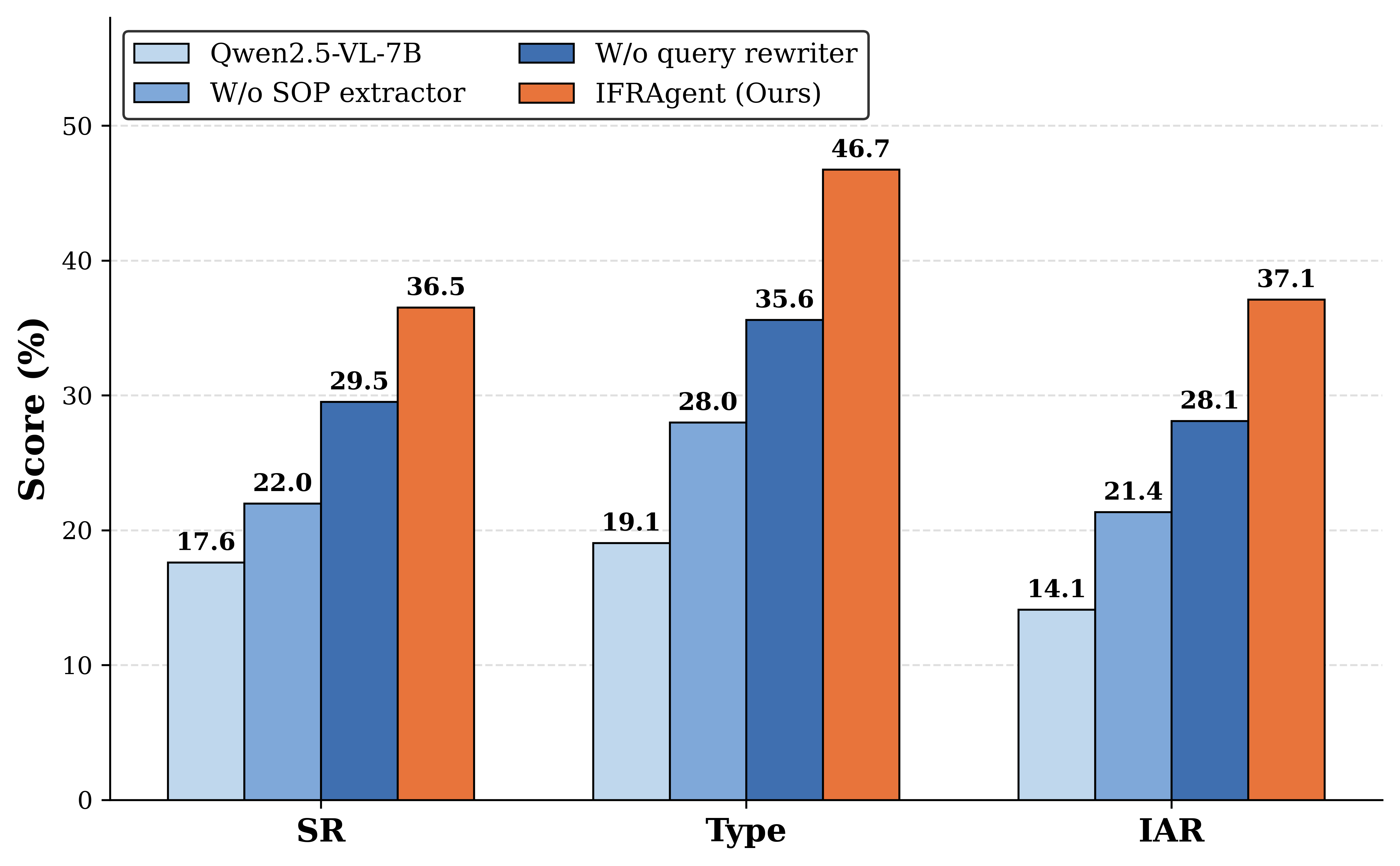}
\caption{Experimental results of ablation study.}
\label{ablation}
\end{figure}
We conducted an ablation study on Qwen2.5-VL-7B to assess the two most critical components of IFRAgent: the SOP extractor $\mathcal{E}$ and the query rewriter $\mathcal{W}$.
As shown in Figure~\ref{ablation}, the query rewriter $\mathcal{W}$ alone brings only a slight improvement, while the SOP extractor $\mathcal{E}$ alone still cannot fully unleash the potential of human demonstrations; combining both components yields the largest gains across all metrics, confirming that both are necessary and complementary.

\begin{figure}[t]
\centering
\includegraphics[width=0.48\textwidth]{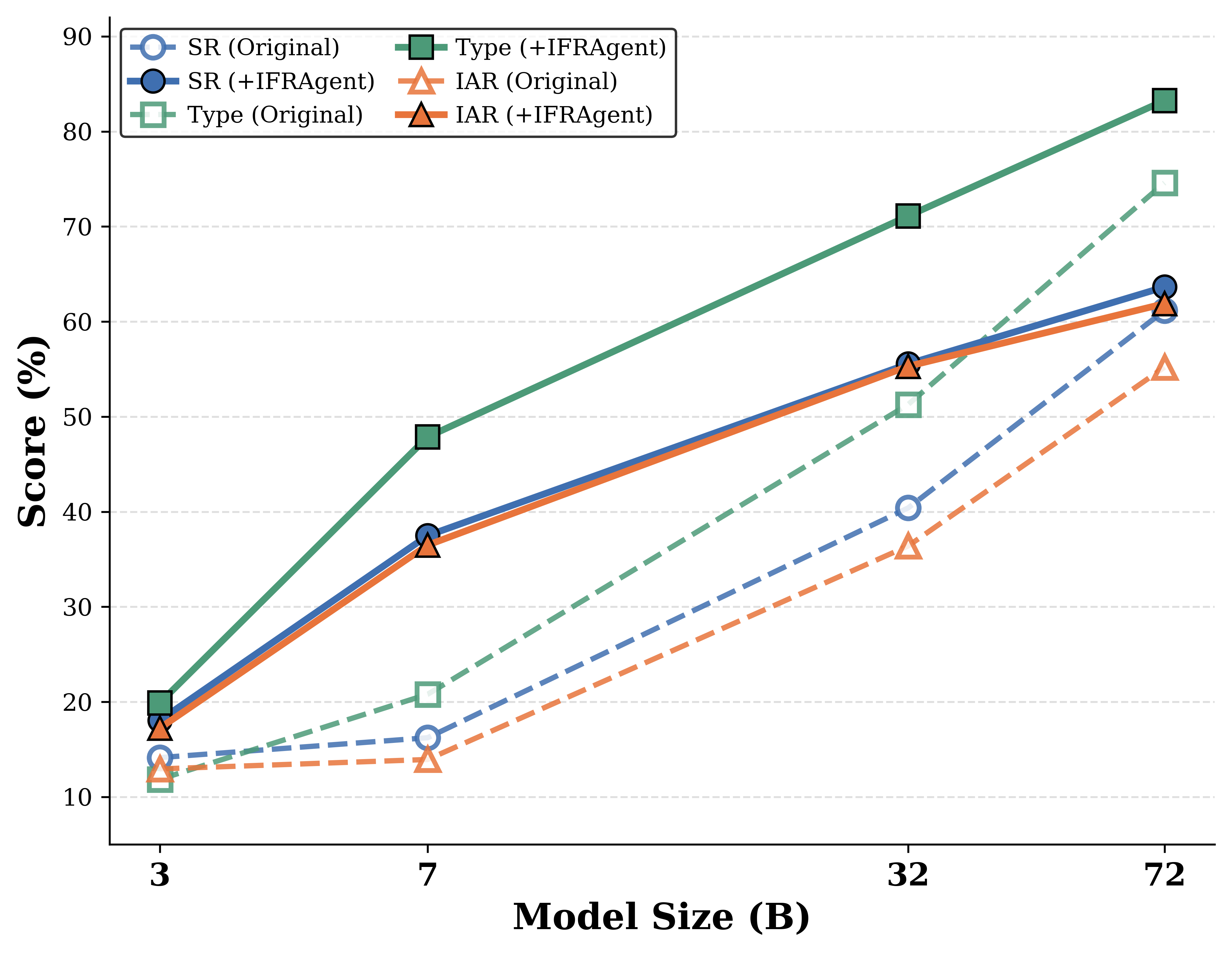}
\caption{IFRAgent on Qwen2.5-VL-3B/7B/32B/72B. Solid/filled markers denote +IFRAgent; dashed/hollow markers denote the original models.}
\label{model_scaling}
\end{figure}

\subsection{Generalizability Analysis on Other Datasets}
\begin{table}[t]
\centering
\begin{tabular}{lp{1cm}p{1.2cm}p{1cm}}
\toprule
\textbf{~~~~~~~~~~~~~Model} & SR(\%)$\uparrow$ & Type(\%)$\uparrow$ & IAR(\%)$\uparrow$ \\
\midrule
OS-Atlas-7B-Pro &58.80 &79.50 &53.80 \\
~~~~~~~~~~~~~~~~~~+ IFRAgent &\textbf{71.80} &\textbf{88.60} &\textbf{71.10} \\
\midrule
UI-TARS-1.5-7B &61.50 &75.90 &53.20 \\
~~~~~~~~~~~~~~~~~~+ IFRAgent &\textbf{65.00} &\textbf{80.00} &\textbf{60.80}\\
\midrule
Qwen2.5-VL-7B &24.30 &27.80 &23.60\\
~~~~~~~~~~~~~~~~~~+ IFRAgent &\textbf{55.40} &\textbf{62.70} &\textbf{54.00}\\
\midrule
GPT-4o+OCR model &40.60& 68.90& 37.70 \\
~~~~~~~~~~~~~~~~~~+ IFRAgent &\textbf{60.20} &\textbf{79.60} &\textbf{56.70}\\
\bottomrule
\end{tabular}
\caption{Generalizability experiment on our modified user-specific OS-Kairos dataset.}
\label{kairos}
\end{table}

To validate the generalizability of IFRAgent, we adapt a portion of the OS-Kairos dataset~\cite{cheng2025kairos}: trajectories from the shop, video, and search scenarios in its training set serve as human demonstrations, while user-intent-aligned actions and ground-truth action lists in the test set are manually annotated following the same protocol as MobileIAR (independent per-user labeling for intent alignment, followed by joint three-annotator labeling for the ground-truth list).

As shown in Table~\ref{kairos}, IFRAgent continues to improve both task completion and human-intent alignment on this modified user-specific OS-Kairos dataset, confirming its generalizability. Consistent with the main results, general-domain models such as Qwen2.5-VL-7B and GPT-4o still outperform specialized mobile-use agent base models like UI-TARS.

%

\subsection{Comparison With Other Methods}

\begin{table}[t]
\centering
\begin{tabular}{lp{1cm}p{1.2cm}p{1cm}}
\toprule
\textbf{~~~Model} & SR(\%)$\uparrow$ & Type(\%)$\uparrow$ & IAR(\%)$\uparrow$ \\
\midrule
OS-Atlas-7B-Pro &41.83 &74.47 &36.58\\
~~~~~~~~~+ LearnAct & 47.99 & 76.19 & 43.65\\
~~~~~~~~~+ SOP demonstration & 50.00 & 81.65 & 41.90\\
~~~~~~~~~+ IFRAgent &\textbf{53.87} &\textbf{85.65} &\textbf{46.85} \\
\midrule
UI-TARS-1.5-7B &43.64 &72.92 &35.10\\
~~~~~~~~~+ LearnAct &42.30 &70.12 &39.57 \\
~~~~~~~~~+ SOP demonstration &39.66 &70.92 &35.82 \\
~~~~~~~~~+ IFRAgent &\textbf{49.24} &\textbf{75.98} &\textbf{46.41} \\
\midrule
Qwen2.5-VL-7B &16.21 &21.54 &13.78 \\
~~~~~~~~~+ LearnAct &20.58 &28.04 &20.82 \\
~~~~~~~~~+ SOP demonstration &23.27 &27.18 &22.58 \\
~~~~~~~~~+ IFRAgent &\textbf{36.69} &\textbf{47.27} &\textbf{34.60} \\
\midrule
GPT-4o+OCR model &37.21 &78.10 &30.76 \\
~~~~~~~~~+ LearnAct & 40.58& 73.89 &36.85 \\
~~~~~~~~~+ SOP demonstration & 44.38& 77.53 &40.09 \\
~~~~~~~~~+ IFRAgent &\textbf{45.26} &\textbf{81.08} &\textbf{41.53} \\
\bottomrule
\end{tabular}
\caption{Comparison with SOP demonstration. IFRAgent better abstracts the user's intention flow, yielding larger gains.}
\label{comparison_method}
\end{table}
To validate IFRAgent's effectiveness in recognizing intention flow from human demonstrations, we compare it against other demonstration-learning baselines on four models. Specifically, we use a demoparser to extract a 1-shot SOP demonstration from user-specific human demonstrations to enhance the prompt, rewrite all queries into ambiguous instructions to enable IAR testing, and additionally compare with the reported results of LearnAct~\cite{liu2025learnact}.

As shown in Table~\ref{comparison_method}, while all demonstration-learning methods improve over the baseline, IFRAgent achieves the largest gains for every model, including GPT-4o.
Although GPT-4o's extensive world knowledge lets it partially abstract implicit intent from the SOP demonstration, this still cannot substitute for IFRAgent's explicit personalized query and SOP rewriting.
The advantage of IFRAgent is even more pronounced for 7B-scale agents, which struggle to abstract implicit intent flow from SOP demonstrations on their own, whereas IFRAgent directly supplies this information via the personalized rewritten query and SOP.

\subsection{Scale Analysis}
\subsubsection{Model Scale Analysis}
To examine how IFRAgent's benefit varies with model capacity, we experiment on Qwen2.5-VL-3B/7B/32B/72B.
As shown in Figure~\ref{model_scaling}, the +IFRAgent curve lies strictly above the Original curve for every metric and every model size, with the largest gains at the 7B and 32B scales.
The 3B model lacks sufficient instruction-following ability to fully exploit the personalized queries and SOPs, while the 72B model already has strong general mobile operation capability, leaving less room for SR improvement (though IAR still improves noticeably).
Overall, IFRAgent benefits moderate-scale mobile-use agents the most.
\begin{figure}[t]
\includegraphics[width=0.45\textwidth]{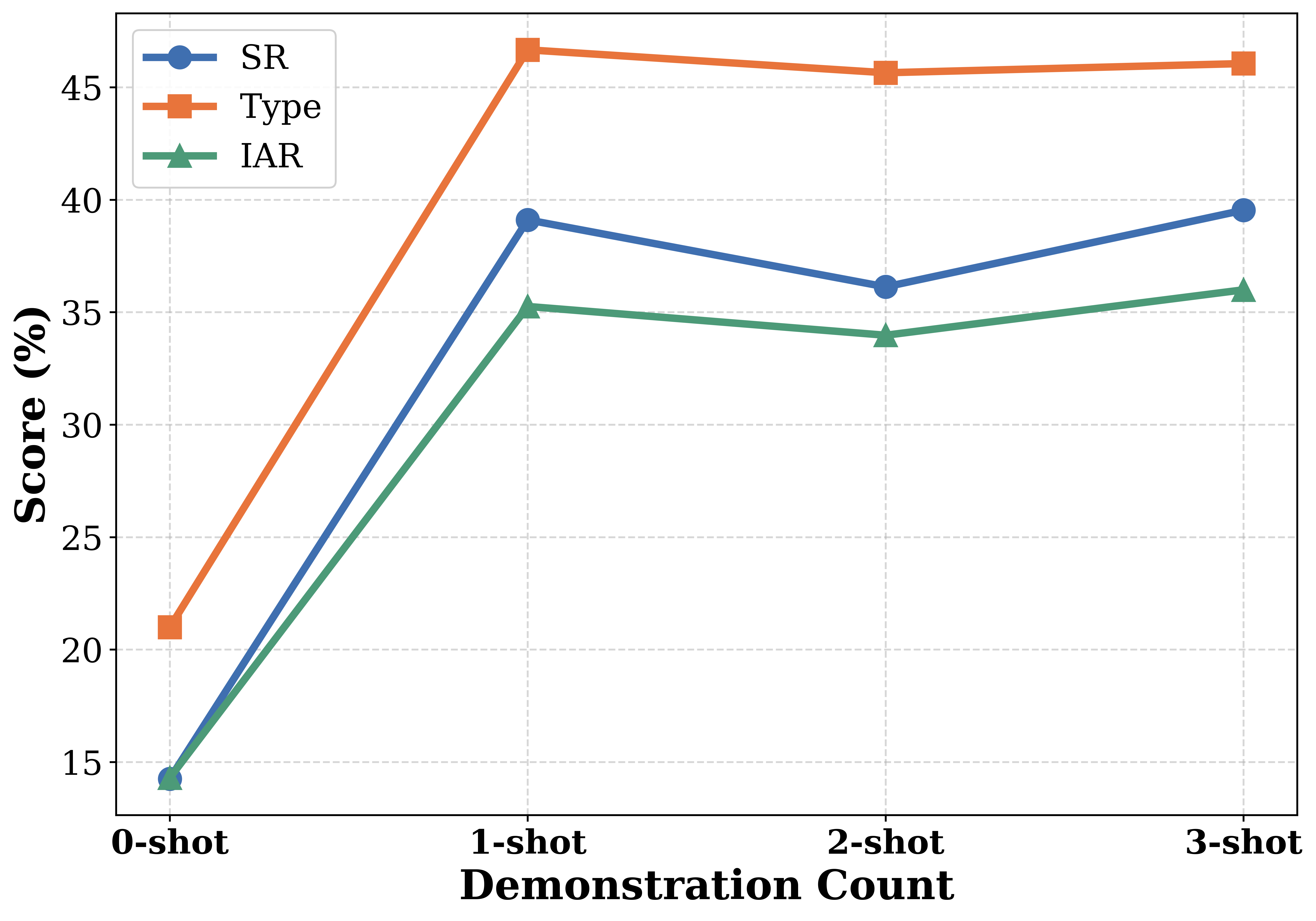}
\caption{Experimental results of IFRAgent-enhanced model under varying numbers of demonstrations.}
\label{kshot}
\end{figure}

\subsubsection{Demonstration Count Analysis}
To study how the number of demonstrations fed to the SOP extractor $\mathcal{E}$ affects performance, we vary the number of $(q', p')$ pairs on Qwen2.5-VL-7B (Figure~\ref{kshot}).
1-shot already yields a large gain over 0-shot, whereas 2-shot and 3-shot bring no further substantial improvement and can even introduce irrelevant information that interferes with $\mathcal{E}$.
Given the added computational cost of more demonstrations, 1-shot SOP extraction is thus a reasonable default for IFRAgent.

\subsection{Time Cost of IFRAgent}
\begin{figure}[t]
\includegraphics[width=0.5\textwidth]{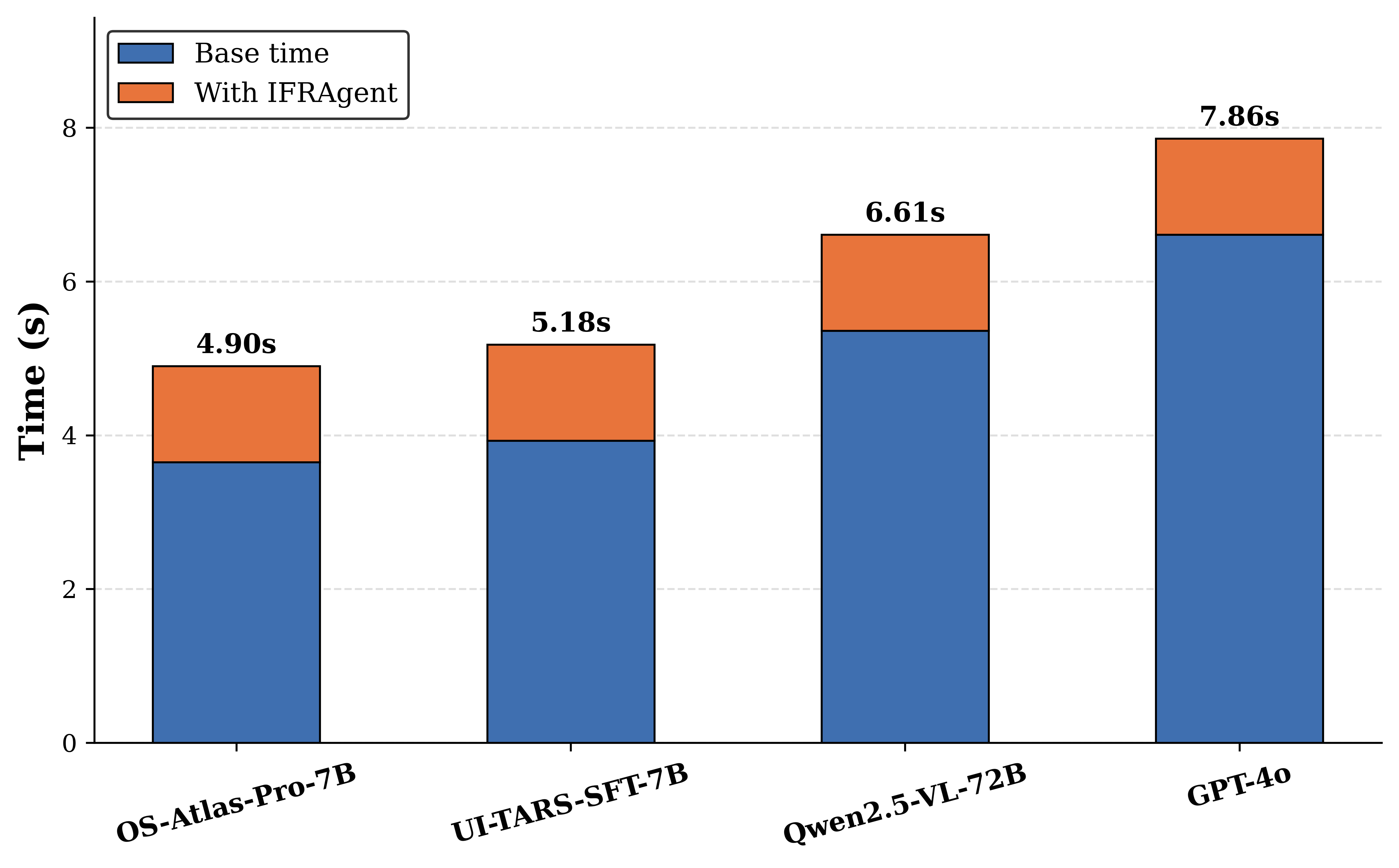}
\caption{Comparison of inference time with and without IFRAgent across different model scales.}
\label{time}
\end{figure}

We measure the time cost of adding IFRAgent on top of base models spanning open-source (7B and 72B) and closed-source models (Figure~\ref{time}).
Since query rewriting and SOP generation are both handled by a single lightweight Qwen3-4B call per query, the additional overhead is minimal, showing that IFRAgent improves human-intent alignment while remaining highly efficient.

\section{Conclusion}
To address the lack of alignment benchmarks and the impracticality of per-user fine-tuning for personalized mobile-use agents, we introduce \textbf{MobileIAR}, a user-specific dataset for assessing intention alignment rate, and \textbf{IFRAgent}, a plug-and-play framework that leverages both explicit and implicit intention flows from human demonstrations to rewrite ambiguous queries into personalized queries and SOPs.
Extensive experiments and further analyses show that IFRAgent improves both task completion and human-intent alignment across diverse mobile-use agents, offering valuable insights for building personalized mobile-use agents.

\bibliography{aaai2027}

@article{zhang2024large,
  title={Large language model-brained gui agents: A survey},
  author={Zhang, Chaoyun and He, Shilin and Qian, Jiaxu and Li, Bowen and Li, Liqun and Qin, Si and Kang, Yu and Ma, Minghua and Lin, Qingwei and Rajmohan, Saravan and others},
  journal={arXiv preprint arXiv:2411.18279},
  year={2024}
}

@inproceedings{wuatlas,
  title={OS-ATLAS: Foundation Action Model for Generalist GUI Agents},
  author={Wu, Zhiyong and Wu, Zhenyu and Xu, Fangzhi and Wang, Yian and Sun, Qiushi and Jia, Chengyou and Cheng, Kanzhi and Ding, Zichen and Chen, Liheng and Liang, Paul Pu and others},
  booktitle={The Thirteenth International Conference on Learning Representations},
  year={2025}
}

@inproceedings{ma2024comprehensive,
    title = "{C}o{C}o-Agent: A Comprehensive Cognitive {MLLM} Agent for Smartphone {GUI} Automation",
    author = "Ma, Xinbei  and
      Zhang, Zhuosheng  and
      Zhao, Hai",
    editor = "Ku, Lun-Wei  and
      Martins, Andre  and
      Srikumar, Vivek",
    booktitle = "Findings of the Association for Computational Linguistics: ACL 2024",
    month = aug,
    year = "2024",
    address = "Bangkok, Thailand",
    publisher = "Association for Computational Linguistics",
    url = "https://aclanthology.org/2024.findings-acl.539/",
    doi = "10.18653/v1/2024.findings-acl.539",
    pages = "9097--9110",
}

@inproceedings{zhang2023you,
    title = "You Only Look at Screens: Multimodal Chain-of-Action Agents",
    author = "Zhang, Zhuosheng  and
      Zhang, Aston",
    editor = "Ku, Lun-Wei  and
      Martins, Andre  and
      Srikumar, Vivek",
    booktitle = "Findings of the Association for Computational Linguistics: ACL 2024",
    month = aug,
    year = "2024",
    address = "Bangkok, Thailand",
    publisher = "Association for Computational Linguistics",
    url = "https://aclanthology.org/2024.findings-acl.186/",
    doi = "10.18653/v1/2024.findings-acl.186",
    pages = "3132--3149",
}

@inproceedings{zhang2024android,
    title = "Android in the Zoo: Chain-of-Action-Thought for {GUI} Agents",
    author = "Zhang, Jiwen  and
      Wu, Jihao  and
      Yihua, Teng  and
      Liao, Minghui  and
      Xu, Nuo  and
      Xiao, Xiao  and
      Wei, Zhongyu  and
      Tang, Duyu",
    editor = "Al-Onaizan, Yaser  and
      Bansal, Mohit  and
      Chen, Yun-Nung",
    booktitle = "Findings of the Association for Computational Linguistics: EMNLP 2024",
    month = nov,
    year = "2024",
    address = "Miami, Florida, USA",
    publisher = "Association for Computational Linguistics",
    url = "https://aclanthology.org/2024.findings-emnlp.702/",
    doi = "10.18653/v1/2024.findings-emnlp.702",
    pages = "12016--12031",
}

@article{liu2025llm,
  title={LLM-Powered GUI Agents in Phone Automation: Surveying Progress and Prospects},
  author={Liu, William and Liu, Liang and Guo, Yaxuan and Xiao, Han and Lin, Weifeng and Chai, Yuxiang and Ren, Shuai and Liang, Xiaoyu and Li, Linghao and Wang, Wenhao and others},
  journal={arXiv preprint arXiv:2412.13501},
  year={2025}
}

@article{wang2024gui,
  title={GUI Agents with Foundation Models: A Comprehensive Survey},
  author={Wang, Shuai and Liu, Weiwen and Chen, Jingxuan and Gan, Weinan and Zeng, Xingshan and Yu, Shuai and Hao, Xinlong and Shao, Kun and Wang, Yasheng and Tang, Ruiming},
  journal={arXiv preprint arXiv:2411.04890},
  year={2024}
}

@inproceedings{wang2024mobile,
  title={Mobile-Agent: Autonomous Multi-Modal Mobile Device Agent with Visual Perception},
  author={Wang, Junyang and Xu, Haiyang and Ye, Jiabo and Yan, Ming and Shen, Weizhou and Zhang, Ji and Huang, Fei and Sang, Jitao},
  booktitle={ICLR 2024 Workshop on Large Language Model (LLM) Agents},
  year={2024}
}

@inproceedings{wang2024distrl,
  title={DistRL: An Asynchronous Distributed Reinforcement Learning Framework for On-Device Control Agent},
  author={Wang, Taiyi and Wu, Zhihao and Liu, Jianheng and Yuen, Derek and Jianye, HAO and Wang, Jun and Shao, Kun},
  booktitle={NeurIPS 2024 Workshop on Fine-Tuning in Modern Machine Learning: Principles and Scalability},
  year={2024}
}

@article{rawles2024androidinthewild,
  title={Androidinthewild: A large-scale dataset for android device control},
  author={Rawles, Christopher and Li, Alice and Rodriguez, Daniel and Riva, Oriana and Lillicrap, Timothy},
  journal={Advances in Neural Information Processing Systems},
  volume={36},
  pages={59708--59728},
  year={2023}
}

@article{qin2025ui,
  title={UI-TARS: Pioneering Automated GUI Interaction with Native Agents},
  author={Qin, Yujia and Ye, Yining and Fang, Junjie and Wang, Haoming and Liang, Shihao and Tian, Shizuo and Zhang, Junda and Li, Jiahao and Li, Yunxin and Huang, Shijue and others},
  journal={arXiv preprint arXiv:2501.12326},
  year={2025}
}

@inproceedings{
rawles2024androidworld,
title={AndroidWorld: A Dynamic Benchmarking Environment for Autonomous Agents},
author={Christopher Rawles and Sarah Clinckemaillie and Yifan Chang and Jonathan Waltz and Gabrielle Lau and Marybeth Fair and Alice Li and William E Bishop and Wei Li and Folawiyo Campbell-Ajala and Daniel Kenji Toyama and Robert James Berry and Divya Tyamagundlu and Timothy P Lillicrap and Oriana Riva},
booktitle={The Thirteenth International Conference on Learning Representations},
year={2025},
url={https://openreview.net/forum?id=il5yUQsrjC}
}

@article{wang2025mobileE,
  title={Mobile-Agent-E: Self-Evolving Mobile Assistant for Complex Tasks},
  author={Wang, Zhenhailong and Xu, Haiyang and Wang, Junyang and Zhang, Xi and Yan, Ming and Zhang, Ji and Huang, Fei and Ji, Heng},
  journal={arXiv preprint arXiv:2501.11733},
  year={2025}
}

@article{wang2025mobile,
  title={Mobile-agent-v2: Mobile device operation assistant with effective navigation via multi-agent collaboration},
  author={Wang, Junyang and Xu, Haiyang and Jia, Haitao and Zhang, Xi and Yan, Ming and Shen, Weizhou and Zhang, Ji and Huang, Fei and Sang, Jitao},
  journal={Advances in Neural Information Processing Systems},
  volume={37},
  pages={2686--2710},
  year={2025}
}

@inproceedings{he2016deep,
  title={Deep residual learning for image recognition},
  author={He, Kaiming and Zhang, Xiangyu and Ren, Shaoqing and Sun, Jian},
  booktitle={Proceedings of the IEEE conference on computer vision and pattern recognition},
  pages={770--778},
  year={2016}
}

@inproceedings{liu2022convnet,
  title={A convnet for the 2020s},
  author={Liu, Zhuang and Mao, Hanzi and Wu, Chao-Yuan and Feichtenhofer, Christoph and Darrell, Trevor and Xie, Saining},
  booktitle={Proceedings of the IEEE/CVF conference on computer vision and pattern recognition},
  pages={11976--11986},
  year={2022}
}

@article{jiang2025appagentx,
  title={AppAgentX: Evolving GUI Agents as Proficient Smartphone Users},
  author={Jiang, Wenjia and Zhuang, Yangyang and Song, Chenxi and Yang, Xu and Zhang, Chi},
  journal={arXiv preprint arXiv:2503.02268},
  year={2025}
}

@article{shaw2023pixels,
  title={From pixels to ui actions: Learning to follow instructions via graphical user interfaces},
  author={Shaw, Peter and Joshi, Mandar and Cohan, James and Berant, Jonathan and Pasupat, Panupong and Hu, Hexiang and Khandelwal, Urvashi and Lee, Kenton and Toutanova, Kristina N},
  journal={Advances in Neural Information Processing Systems},
  volume={36},
  pages={34354--34370},
  year={2023}
}

@article{cheng2025kairos,
  title={OS-Kairos: Adaptive Interaction for MLLM-Powered GUI Agents},
  author={Cheng, Pengzhou and Wu, Zheng and Wu, Zongru and Zhang, Aston and Zhang, Zhuosheng and Liu, Gongshen},
  journal={arXiv preprint arXiv:2503.16465},
  year={2025}
}

@inproceedings{lu2025ui,
  title={Ui-r1: Enhancing efficient action prediction of gui agents by reinforcement learning},
  author={Lu, Zhengxi and Chai, Yuxiang and Guo, Yaxuan and Yin, Xi and Liu, Liang and Wang, Hao and Xiao, Han and Ren, Shuai and Zhao, Pengxiang and Liu, Guangyi and others},
  booktitle={Proceedings of the AAAI Conference on Artificial Intelligence},
  volume={40},
  number={21},
  pages={17608--17616},
  year={2026}
}

@article{verma2024adaptagent,
  title={Adaptagent: Adapting multimodal web agents with few-shot learning from human demonstrations},
  author={Verma, Gaurav and Kaur, Rachneet and Srishankar, Nishan and Zeng, Zhen and Balch, Tucker and Veloso, Manuela},
  journal={arXiv preprint arXiv:2411.13451},
  year={2024}
}

@article{liu2025learnact,
  title={Learnact: Few-shot mobile gui agent with a unified demonstration benchmark},
  author={Liu, Guangyi and Zhao, Pengxiang and Liu, Liang and Chen, Zhiming and Chai, Yuxiang and Ren, Shuai and Wang, Hao and He, Shibo and Meng, Wenchao},
  journal={arXiv preprint arXiv:2504.13805},
  year={2025}
}

@article{correia2024survey,
  title={A survey of demonstration learning},
  author={Correia, Andre and Alexandre, Luis A},
  journal={Robotics and Autonomous Systems},
  volume={182},
  pages={104812},
  year={2024},
  publisher={Elsevier}
}

@inproceedings{rybski2007interactive,
  title={Interactive robot task training through dialog and demonstration},
  author={Rybski, Paul E and Yoon, Kevin and Stolarz, Jeremy and Veloso, Manuela M},
  booktitle={Proceedings of the ACM/IEEE international conference on Human-robot interaction},
  pages={49--56},
  year={2007}
}

@inproceedings{ng2000algorithms,
  title={Algorithms for inverse reinforcement learning.},
  author={Ng, Andrew Y and Russell, Stuart and others},
  booktitle={Icml},
  volume={1},
  pages={2},
  year={2000}
}

@article{argall2009survey,
  title={A survey of robot learning from demonstration},
  author={Argall, Brenna D and Chernova, Sonia and Veloso, Manuela and Browning, Brett},
  journal={Robotics and autonomous systems},
  volume={57},
  number={5},
  pages={469--483},
  year={2009},
  publisher={Elsevier}
}

@inproceedings{zhang-etal-2024-mm,
    title = "{MM}-{LLM}s: Recent Advances in {M}ulti{M}odal Large Language Models",
    author = "Zhang, Duzhen  and
      Yu, Yahan  and
      Dong, Jiahua  and
      Li, Chenxing  and
      Su, Dan  and
      Chu, Chenhui  and
      Yu, Dong",
    editor = "Ku, Lun-Wei  and
      Martins, Andre  and
      Srikumar, Vivek",
    booktitle = "Findings of the Association for Computational Linguistics: ACL 2024",
    month = aug,
    year = "2024",
    address = "Bangkok, Thailand",
    publisher = "Association for Computational Linguistics",
    url = "https://aclanthology.org/2024.findings-acl.738/",
    doi = "10.18653/v1/2024.findings-acl.738",
    pages = "12401--12430"
}

@article{yin2024survey,
  title={A survey on multimodal large language models},
  author={Yin, Shukang and Fu, Chaoyou and Zhao, Sirui and Li, Ke and Sun, Xing and Xu, Tong and Chen, Enhong},
  journal={National Science Review},
  volume={11},
  number={12},
  pages={nwae403},
  year={2024},
  publisher={Oxford University Press}
}

@article{Qwen2.5-VL,
  title={Qwen2.5-VL Technical Report},
  author={Bai, Shuai and Chen, Keqin and Liu, Xuejing and Wang, Jialin and Ge, Wenbin and Song, Sibo and Dang, Kai and Wang, Peng and Wang, Shijie and Tang, Jun and Zhong, Humen and Zhu, Yuanzhi and Yang, Mingkun and Li, Zhaohai and Wan, Jianqiang and Wang, Pengfei and Ding, Wei and Fu, Zheren and Xu, Yiheng and Ye, Jiabo and Zhang, Xi and Xie, Tianbao and Cheng, Zesen and Zhang, Hang and Yang, Zhibo and Xu, Haiyang and Lin, Junyang},
  journal={arXiv preprint arXiv:2502.13923},
  year={2025}
}

@article{Qwen-VL,
  title={Qwen-VL: A Versatile Vision-Language Model for Understanding, Localization, Text Reading, and Beyond},
  author={Bai, Jinze and Bai, Shuai and Yang, Shusheng and Wang, Shijie and Tan, Sinan and Wang, Peng and Lin, Junyang and Zhou, Chang and Zhou, Jingren},
  journal={arXiv preprint arXiv:2308.12966},
  year={2023}
}

@misc{glm2024chatglm,
      title={ChatGLM: A Family of Large Language Models from GLM-130B to GLM-4 All Tools},
      author={Team GLM and Aohan Zeng and Bin Xu and Bowen Wang and Chenhui Zhang and Da Yin and Diego Rojas and Guanyu Feng and Hanlin Zhao and Hanyu Lai and Hao Yu and Hongning Wang and Jiadai Sun and Jiajie Zhang and Jiale Cheng and Jiayi Gui and Jie Tang and Jing Zhang and Juanzi Li and Lei Zhao and Lindong Wu and Lucen Zhong and Mingdao Liu and Minlie Huang and Peng Zhang and Qinkai Zheng and Rui Lu and Shuaiqi Duan and Shudan Zhang and Shulin Cao and Shuxun Yang and Weng Lam Tam and Wenyi Zhao and Xiao Liu and Xiao Xia and Xiaohan Zhang and Xiaotao Gu and Xin Lv and Xinghan Liu and Xinyi Liu and Xinyue Yang and Xixuan Song and Xunkai Zhang and Yifan An and Yifan Xu and Yilin Niu and Yuantao Yang and Yueyan Li and Yushi Bai and Yuxiao Dong and Zehan Qi and Zhaoyu Wang and Zhen Yang and Zhengxiao Du and Zhenyu Hou and Zihan Wang},
      year={2024},
      eprint={2406.12793},
      archivePrefix={arXiv}
}

@misc{openai2023gpt4,
  title        = {GPT-4: An artificial intelligence model},
  author       = {OpenAI},
  year         = {2023},
  url          = {https://openai.com/research/gpt-4},
}

@article{cheng2025navi,
  title={Navi-plus: Managing Ambiguous GUI Navigation Tasks with Follow-up},
  author={Cheng, Ziming and Huang, Zhiyuan and Pan, Junting and Hou, Zhaohui and Zhan, Mingjie},
  journal={arXiv preprint arXiv:2503.24180},
  year={2025}
}

@article{xu2024androidlab,
  title={Androidlab: Training and systematic benchmarking of android autonomous agents},
  author={Xu, Yifan and Liu, Xiao and Sun, Xueqiao and Cheng, Siyi and Yu, Hao and Lai, Hanyu and Zhang, Shudan and Zhang, Dan and Tang, Jie and Dong, Yuxiao},
  journal={arXiv preprint arXiv:2410.24024},
  year={2024}
}

@article{lewis2020retrieval,
  title={Retrieval-augmented generation for knowledge-intensive nlp tasks},
  author={Lewis, Patrick and Perez, Ethan and Piktus, Aleksandra and Petroni, Fabio and Karpukhin, Vladimir and Goyal, Naman and K{\"u}ttler, Heinrich and Lewis, Mike and Yih, Wen-tau and Rockt{\"a}schel, Tim and others},
  journal={Advances in neural information processing systems},
  volume={33},
  pages={9459--9474},
  year={2020}
}

@inproceedings{hu-etal-2025-os,
    title = "{OS} Agents: A Survey on {MLLM}-based Agents for Computer, Phone and Browser Use",
    author = "Hu, Xueyu  and
      Xiong, Tao  and
      Yi, Biao  and
      Wei, Zishu  and
      Xiao, Ruixuan  and
      Chen, Yurun  and
      Ye, Jiasheng  and
      Tao, Meiling  and
      Zhou, Xiangxin  and
      Zhao, Ziyu  and
      Li, Yuhuai  and
      Xu, Shengze  and
      Wang, Shenzhi  and
      Xu, Xinchen  and
      Qiao, Shuofei  and
      Wang, Zhaokai  and
      Kuang, Kun  and
      Zeng, Tieyong  and
      Wang, Liang  and
      Li, Jiwei  and
      Jiang, Yuchen Eleanor  and
      Zhou, Wangchunshu  and
      Wang, Guoyin  and
      Yin, Keting  and
      Zhao, Zhou  and
      Yang, Hongxia  and
      Wu, Fan  and
      Zhang, Shengyu  and
      Wu, Fei",
    editor = "Che, Wanxiang  and
      Nabende, Joyce  and
      Shutova, Ekaterina  and
      Pilehvar, Mohammad Taher",
    booktitle = "Proceedings of the 63rd Annual Meeting of the Association for Computational Linguistics (Volume 1: Long Papers)",
    month = jul,
    year = "2025",
    address = "Vienna, Austria",
    publisher = "Association for Computational Linguistics",
    url = "https://aclanthology.org/2025.acl-long.369/",
    pages = "7436--7465",
    ISBN = "979-8-89176-251-0"
}

@article{li2025mobileuse,
  title={MobileUse: A GUI Agent with Hierarchical Reflection for Autonomous Mobile Operation},
  author={Li, Ning and Qu, Xiangmou and Zhou, Jiamu and Wang, Jun and Wen, Muning and Du, Kounianhua and Lou, Xingyu and Peng, Qiuying and Zhang, Weinan},
  journal={arXiv preprint arXiv:2507.16853},
  year={2025}
}

@misc{tang2025surveymllmbasedguiagents,
      title={A Survey on (M)LLM-Based GUI Agents}, 
      author={Fei Tang and Haolei Xu and Hang Zhang and Siqi Chen and Xingyu Wu and Yongliang Shen and Wenqi Zhang and Guiyang Hou and Zeqi Tan and Yuchen Yan and Kaitao Song and Jian Shao and Weiming Lu and Jun Xiao and Yueting Zhuang},
      year={2025},
      eprint={2504.13865},
      archivePrefix={arXiv},
      primaryClass={cs.HC},
      url={https://arxiv.org/abs/2504.13865}, 
}

@article{ye2025mobile,
  title={Mobile-agent-v3: Fundamental agents for gui automation},
  author={Ye, Jiabo and Zhang, Xi and Xu, Haiyang and Liu, Haowei and Wang, Junyang and Zhu, Zhaoqing and Zheng, Ziwei and Gao, Feiyu and Cao, Junjie and Lu, Zhengxi and others},
  journal={arXiv preprint arXiv:2508.15144},
  year={2025}
}

@article{Qwen3-VL,
      title={Qwen3-VL Technical Report}, 
      author={Shuai Bai and Yuxuan Cai and Ruizhe Chen and Keqin Chen and Xionghui Chen and Zesen Cheng and Lianghao Deng and Wei Ding and Chang Gao and Chunjiang Ge and Wenbin Ge and Zhifang Guo and Qidong Huang and Jie Huang and Fei Huang and Binyuan Hui and Shutong Jiang and Zhaohai Li and Mingsheng Li and Mei Li and Kaixin Li and Zicheng Lin and Junyang Lin and Xuejing Liu and Jiawei Liu and Chenglong Liu and Yang Liu and Dayiheng Liu and Shixuan Liu and Dunjie Lu and Ruilin Luo and Chenxu Lv and Rui Men and Lingchen Meng and Xuancheng Ren and Xingzhang Ren and Sibo Song and Yuchong Sun and Jun Tang and Jianhong Tu and Jianqiang Wan and Peng Wang and Pengfei Wang and Qiuyue Wang and Yuxuan Wang and Tianbao Xie and Yiheng Xu and Haiyang Xu and Jin Xu and Zhibo Yang and Mingkun Yang and Jianxin Yang and An Yang and Bowen Yu and Fei Zhang and Hang Zhang and Xi Zhang and Bo Zheng and Humen Zhong and Jingren Zhou and Fan Zhou and Jing Zhou and Yuanzhi Zhu and Ke Zhu},
	  journal={arXiv preprint arXiv:2511.21631},
      year={2025}
}

@inproceedings{zhang2026tongui,
  title={Tongui: Internet-scale trajectories from multimodal web tutorials for generalized gui agents},
  author={Zhang, Bofei and Shang, Zirui and Gao, Zhi and Zhang, Wang and Xie, Rui and Ma, Xiaojian and Yuan, Tao and Wu, Xinxiao and Zhu, Song-Chun and Li, Qing},
  booktitle={Proceedings of the AAAI Conference on Artificial Intelligence},
  volume={40},
  number={15},
  pages={12367--12375},
  year={2026}
}

@inproceedings{jung2026tvagent,
  title={TVAgent: A lightweight Vision-Language-Model for TV GUI Agent},
  author={Jung, Hoin and Liu, Jinyang and Rao, Anirudh and Kim, Honghoe and Zhao, Xiangyuan and Chandra, Ashwin and Sarkis, Michel},
  booktitle={4th Deployable AI Workshop},
  year={2026}
}

@article{chen2026trace,
  title={TRACE: Trajectory-Aware Comprehensive Evaluation for Deep Research Agents},
  author={Chen, Yanyu and Jiang, Jiyue and Liu, Jiahong and Zhang, Yifei and Guo, Xiao and King, Irwin},
  journal={arXiv preprint arXiv:2602.21230},
  year={2026}
}


\end{document}